\documentclass{article}

\usepackage[final]{neurips_2021}

\usepackage[utf8]{inputenc} 
\usepackage[T1]{fontenc}    
\usepackage{hyperref}       
\usepackage{url}            
\usepackage{booktabs}       
\usepackage{amsfonts}       
\usepackage{nicefrac}       
\usepackage{microtype}      
\usepackage{xcolor}         
\usepackage{graphicx}
\usepackage{natbib}
\usepackage{hyperref}
\usepackage{subfigure}
\usepackage{multicol}
\setcitestyle{square,numbers}
\title{Evaluating Predictive Uncertainty and Robustness to Distributional Shift Using Real World Data}
\author{
  Kumud Lakara  \thanks{Equal contribution.}\\
  Dept. of Computer Science and Engineering\\
  Manipal Institute of Technology, MAHE \\
  Manipal, India\\
  \texttt{lakara.kumud@gmail.com} \\
   \And
   Akshat Bhandari \footnotemark[1] \\
  Dept. of Computer Science and Engineering\\
  Manipal Institute of Technology, MAHE \\
  Manipal, India\\
   \texttt{akshatbhandari15@gmail.com}
   \And
   Pratinav Seth \footnotemark[1]\\
   Dept. of Computer Applications \\
   Manipal Institute of Technology, MAHE \\
   Manipal, India \\
   \texttt{seth.pratinav@gmail.com} \\
  \And
   Ujjwal Verma \\
  Dept. of Electronics and Comm. Engineering \\
  Manipal Institute of Technology, MAHE \\
  Manipal, India \\
  \texttt{ujjwal.verma@manipal.edu} \\
}

\begin{document}

\maketitle

\begin{abstract}
Most machine learning models operate under the assumption that the training, testing and deployment data is independent and identically distributed (i.i.d.). This assumption does not generally hold true in a natural setting. Usually, the deployment data is subject to various types of distributional shifts. The magnitude of a model's performance is proportional to this shift in the distribution of the dataset. Thus it becomes necessary to evaluate a model's uncertainty and robustness to distributional shifts to get a realistic estimate of its expected performance on real-world data. Present methods to evaluate uncertainty and model's robustness are lacking and often fail to paint the full picture. Moreover, most analysis so far has primarily focused on classification tasks. In this paper, we propose more insightful metrics for general regression tasks using the Shifts Weather Prediction Dataset. We also present an evaluation of the baseline methods using these metrics.
\end{abstract}

\section{Introduction}
\label{intro}
Recent times have seen the growing deployment of machine learning models and deep neural networks in many mission-critical tasks such as medical image diagnosis \cite{ker2017deep}, banking systems \cite{boobier2020ai} and autonomous vehicles \cite{nascimento2018concerns}. A prevalent assumption in machine learning is that the data the model sees after the development phase is independent and identically distributed (i.i.d). This implies that if a model performs well during the development phase it will be able to interpolate that performance to deployment. This however is seldom the case. 
Most data available to models in the real world is often out of the distribution of the data on which the models were trained and tested. When dealing with critical applications, the model’s inability to generalize to the out of distribution deployment data can have disastrous consequences. Therefore, such tasks require not only for models to make accurate predictions but also a precise quantification of predictive uncertainty \cite{amodei2016concrete}. Thus, it is imperative to properly assess a model’s robustness to distribution shift and its estimation of predictive uncertainty.  

While the problem is apparent, yet there are limited resources and datasets that allow for proper evaluation of uncertainty estimates and robustness to distributional shift emulating real-world data. Most of the available datasets such as Imagenet - C \cite{hendrycks2019benchmarking}, A \cite{hendrycks2021natural}, R \cite{hendrycks2020many}, O \cite{hendrycks2021natural} and WILDS \cite{koh2021wilds} focus primarily on image classification tasks. The recently introduced Shifts Dataset \cite{malinin2021shifts} provides a favourable data setting. It is composed of three parts each corresponding to a different data modality: tabular weather prediction data, machine translation data and self-driving car data. All the data modalities are affected by distributional shift and pose challenges with respect to evaluating uncertainty predictions and robustness against distributional shift. 

In this work, we present novel evaluation metrics for predictions on the tabular data related to weather prediction in the Shifts Dataset \cite{malinin2021shifts}. We introduce metrics for effectively evaluating both uncertainty predictions \textit{and} robustness to distributional shift for regression tasks. We believe these metrics will be a much needed addition to the scarce pool of metrics available presently for regression tasks. In addition to this, we also validate the proposed metrics by evaluating the performance of baselines for predicting the temperature at a particular latitude/longitude and time, given all available measurements and climate model predictions.\footnote{ \url{https://github.com/kumudlakara/Shifts_evaluation_metrics_regression}}

\section{Evaluation Metrics}
 We believe a model with lower degradation in performance on shifted data is more robust to distribution shift. The quality of predictive uncertainty is governed by the model's ability to discriminate between i.i.d. and out of distribution data. We present metrics for the joint evaluation of predictive uncertainty and robustness to distributional shift. We validate our proposed metrics using the baseline Gradient Boosted Decision Trees (GBDT) models as used in \cite{malinin2021shifts}. 

To build baseline gradient boosted decision trees for the temperature prediction regression task, the open-source
CatBoost gradient boosting library \cite{prokhorenkova2017catboost} is used. Similar to Malinin et. al \cite{malinin2021shifts}, we consider an ensemble-based approach to uncertainty estimation for GBDT models \cite{Ustimenko2021UncertaintyIG}. We use a pre-trained ensemble of ten models on the train data from the canonical partition of
the Weather Prediction dataset with different random seeds. The models are optimized with the loss function \verb|RMSEWithUncertainty| \cite{Ustimenko2021UncertaintyIG} which predicts mean and variance of
the normal distribution by optimizing the negative log-likelihood. Hyperparameter tuning is performed on the in-domain development data. Each model is constructed with a
depth of 8 and is then trained for 20,000 iterations with a learning rate of 0.3.

\subsection{RMSE/RMV Ratio}
The root mean squared error (RMSE) and the root mean variance (RMV) can be viewed as indicators of the model accuracy and uncertainty respectively. 
\begin{multicols}{2}
\noindent
\begin{equation}
\label{eqn:rmseEqn}
    RMSE =  \sqrt{ \frac{1}{|B|}\sum_{b \epsilon B} (y_b - \hat{y_b})^2}
\end{equation}
\begin{equation}
\label{eqn:rmvEqn}
    RMV = \sqrt{ \frac{1}{|B|}\sum_{b \epsilon B} \sigma_b^2}
\end{equation}
\end{multicols}
where $\sigma_b^2$ is the variance, $|B|$ is the bin size and $y_b - \hat{y_b}$ is the error between the model's prediction and the ground truth value for example b $\epsilon$ B.
The ratio of the two should be as close to one as possible. We reason this inference by considering the following paradigm: 

\emph{If the error (RMSE) for a prediction is high then the model's uncertainty (RMV) about that prediction should also be high.}

\begin{figure}
    \centering
    \label{fig:figure1}
    \subfigure[]{\includegraphics[width=0.45\textwidth]{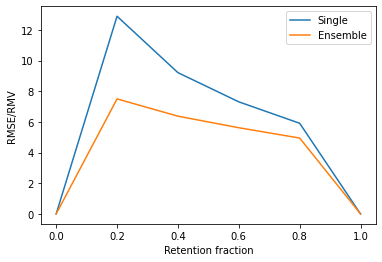}}
    \includegraphics[scale=0.45]{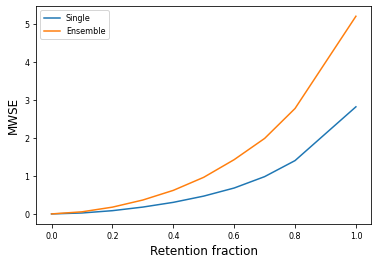}
    \caption{(a) RMSE/RMV retention curve (b) MWSE retention curve}
    
\end{figure}

When plotting the retention curve \cite{malinin2019uncertainty}\cite{lakshminarayanan2016simple} for the RMSE/RMV ratio, an ideal scenario would be a horizontal line close to one. This would indicate that RMSE is roughly equal to RMV.
 \hyperref[fig:figure1]{Figure 1(a)} shows the RMSE/RMV retention curves (R3-curves) for single and ensemble models. The ensemble model clearly displays a more linear relationship between RMSE and RMV compared to the single model. The RMSE/RMV ratio is also much closer to one for the ensemble model compared to the single model.

\subsection{Logarithmic Expected Normalized Calibration Error (LENCE)}
 Levi et al. \cite{levi2019evaluating} propose a definition for calibration for regression by replacing mis-classification probability with mean squared error. This is called Expected Normalized Calibration Error (ENCE). This is an improvement over the existing definition of calibrated regression uncertainty \cite{kuleshov2018accurate}.
To calculate ENCE we arrange the data in increasing order of variance. Then we create bins of size S such that S divides the total number of examples into N bins. Each resulting bin represents an interval in the standard deviation axis. These intervals are non-overlapping and their boundary values are in  increasing order \cite{levi2019evaluating}. This measure is now analogous to the Expected Calibration Error (ECE)\cite{naeini2015obtaining} used in the case of classification problems. 
For finding the error in calibration of the model, we use the following formulation for ENCE:
\begin{equation}
\label{eqn:ence}
    ENCE = \frac{1}{N}\sum_{b=1}^{N}\frac{|RMV(b) - RMSE(b)|}{RMV(b)}
\end{equation}
where $N$ is the total number of bins. 
A shortcoming of ENCE is that if the model were to predict a homogeneous uncertainty for each example and that happened to match the model's empirical uncertainty for the entire population then ENCE would be zero. However, it would not be a correct evaluation of the models uncertainty or generalization ability. Levi et. al \cite{levi2019evaluating} propose using the Variation Coefficient of Standard Deviation, $C_v$ as a  secondary diagnostic tool with ENCE.
\begin{equation}
\label{eqn:cv}
    C_v = \frac{\sqrt{\frac{\sum_{t=1}^{T}(\sigma_t - \mu_\sigma)^2}{T-1}}}{\mu_\sigma}
\end{equation}
where $\mu_\sigma = \frac{1}{T}\sum_{t=1}^{T}\sigma_t$ and T is the number of training examples.
Evaluating two separate metrics and drawing relative inferences is complex and hence we propose the Logarithmic ENCE (LENCE). We incorporate $C_v$ as a measure of dispersion of the uncertainty predictions by the model and ENCE as a measure of the calibration error into a single evaluation metric. We define LENCE as: 
\begin{equation}
    LENCE = log|ENCE + \frac{1}{C_v}|
\end{equation}
Ideally, the uncertainty values will be well dispersed and $C_v$ would be high. Then LENCE will approximately be $log|ENCE|$. In the absolute worst case when $ENCE=0$ and $C_v\rightarrow0$; $LENCE\rightarrow \infty$. ENCE, $C_v$ and LENCE values for single and ensemble models are presented in Table \ref{table:ence_cv_tab}. A lower LENCE value for the ensemble model indicates its superiority.

\begin{table}
\caption{ENCE, $C_v$ and LENCE for baseline GBDT ensemble and single models}
\begin{center}
\begin{tabular}{cccc}
\toprule
Model & ENCE & $C_v$ & LENCE \\ \midrule
Ensemble & 1.084   & 0.387  & 1.2968  \\
Single   & 3.551   & 0.550  & 1.6804 \\ \bottomrule
\end{tabular}
\end{center}
\label{table:ence_cv_tab}
\end{table}

\subsection{LL-Fisher Uncertainty (LL-FU)}
\label{llfu}
We introduce the LL-Fisher metric as a joint measure of uncertainty and robustness to distributional shift. The LL-Fisher metric measures the localisation of a probability distribution function \cite{villani2002review}. This metric can be used to evaluate how well a prediction represents the distribution of data that the model used to make the prediction itself. Inspired by log-loss used for calculating Fisher Information \cite{ly2017tutorial} for a prediction \emph{x} with mean {\textmu} and variance {$\sigma^2$} following a continuous distribution. 

 \begin{figure}
    \centering
    \label{fig:figure2}
    \subfigure[]{\includegraphics[width=0.45\textwidth]{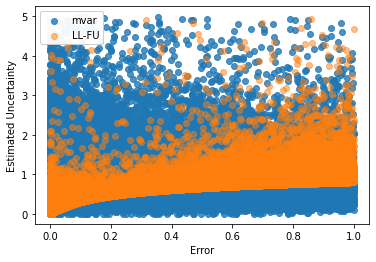}}
    \subfigure[]{\includegraphics[width=0.45\textwidth]{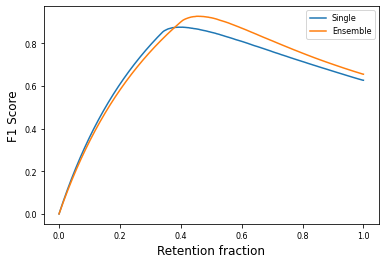}}
    \subfigure[]{\includegraphics[width=0.45\textwidth]{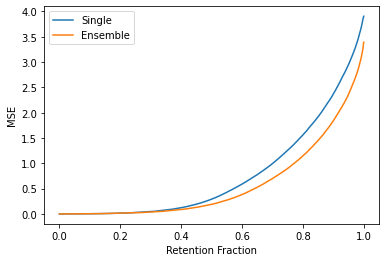}}
    \caption{(a) LL-FU vs error scatter plot (b) LL-FU F1 retention curve (c) MSE - LLFU retention curve}
\end{figure}

We propose the LL-Fisher Uncertainty as follows:
\begin{equation}
\label{sec:llfu}
    LL\scalebox{0.80}[1.0]{-}FU = max(0,\frac{1}{2}log(2\pi\sigma^2)) + \frac{(x-\mu)^2}{2\sigma^2}
\end{equation}
We plot the F1-retention curve \hyperref[fig:figure2]{Figure 2(b)}, the MSE-retention curve \hyperref[fig:figure2]{Figure 2(c)} and calculate the summary statistics: F1-AUC and R-AUC for the evaluation of our metric. We use F1@95 score to jointly evaluate uncertainty and robustness. A good uncertainty measure should achieve low R-AUC, high F1-AUC and high F1@95 scores \cite{malinin2021shifts}. 
We calculated these metrics for various uncertainty measures of prediction using a GBDT model as well as an ensemble of these models. 

\begin{table}[]
 \caption{Ensemble model F1-AUC, F1@95 and R-AUC scores for uncertainty measures on the dev
set of the Shifts Weather Prediction Dataset}
    \centering
    \begin{tabular}{cccccc}
         \toprule
\multicolumn{1}{l}{} & \multicolumn{5}{c}{Uncertainty Metric}                                   \\ \cmidrule{2-6} 
                     & mvar         & tvar         & varm         & epkl         & LL-FU         \\ \midrule
F1-AUC               & 0.4951   & 0.5220 & 0.5012 & 0.5051 & \textbf{0.7009}  \\
F1@95                & 0.6581 & 0.6583 & 0.6579 & 0.6551 & \textbf{0.6778} \\
R-AUC                & 1.4186 & 1.2708 & 1.2868 & 1.2266 & \textbf{0.5838} \\\bottomrule
    \end{tabular}
    \label{table:metrics}
\end{table}

\begin{table}[hbt!]
 \caption{Single model F1-AUC, F1@95 and R-AUC scores for uncertainty measures on the dev set of the Shifts Weather Prediction Dataset.}
    \centering
    \begin{tabular}{cccccc}
         \toprule
\multicolumn{1}{l}{} & \multicolumn{5}{c}{Uncertainty Metric}                                   \\ \cmidrule{2-6} 
                     & mvar         & tvar         & varm         & epkl         & LL-FU         \\ \midrule
F1-AUC               & 0.4425 & 0.4425 & 0.3851 & 0.4425 & \textbf{0.6873}  \\
F1@95                & 0.6276 & 0.6276 & 0.6115 & 0.6276 & \textbf{0.6478} \\
R-AUC                & 1.8868 & 1.8868 & 1.9526 & 1.8868 & \textbf{0.7740} \\\bottomrule
    \end{tabular}
    \label{table:llfu_sin}
\end{table}

Results using ensemble and single models for various uncertainty measures are depicted in Table \ref{table:metrics} and Table \ref{table:llfu_sin} showing superiority of our uncertainty measure. We compare both scores for ensemble and single model for LLFU. The scores for ensemble are better than the single model indicating the superiority of the ensembling approach. These results are presented in Table \ref{table:llfu_ens_sin} . 

We observe that for out of distribution data, as the error of prediction increases, the lower bound of uncertainty (LL-FU) also increases. This validates the idea that with increasing shift in data distribution, not only does the likelihood of an error increase but so does the uncertainty corresponding to the prediction. We believe that LL-FU is unique in capturing this property when compared to other uncertainty metrics \cite{malinin2021shifts} as reflected in \hyperref[fig:figure1]{Figure 2(a)}.

\begin{table}[hbt!]
\caption{LL-FU for baseline GBDT ensemble and single models}
\begin{center}
\begin{tabular}{cccc}
\toprule
Model & F1-AUC & F1@95 & R-AUC \\ \midrule
Ensemble & \textbf{0.7009}   & \textbf{0.6778}  & \textbf{0.5838}  \\
Single   & 0.6873   & 0.6478  & 0.7740 \\ \bottomrule
\end{tabular}
\end{center}
\label{table:llfu_ens_sin}
\end{table}

\subsection{Mean Weighted Squared Error (MWSE)}
Intuitively, a model should be increasingly uncertain about it's prediction as the magnitude of distributional shift increases and hence will make more erroneous decisions. The error-retention curves \cite{malinin2019uncertainty} \cite{lakshminarayanan2016simple} do not directly evaluate the uncertainty estimates of the model and only take into account the correlation between error and uncertainty estimates. To more resiliently evaluate robustness and also take into account the magnitude of estimated uncertainty, we introduce MWSE. We propose MWSE as a secondary metric only to gain more insight into the models' behaviour by analysing their relative divergence. 
\begin{equation}
\label{mwseEqn}
    MWSE = \frac{1}{N}\sum_{i=1}^{i=N}(\hat{y_i} - y)^2*(UNC)_i
\end{equation}

where \emph{$\hat{y_i}$} and
\textit{y} is the predicted and true value respectively and \textit{UNC} is model's uncertainty estimate for each prediction.     

\label{headings}

Ideally datapoints with low MSE should have low uncertainty and those with high MSE should have high uncertainty which would mean their product would be very low or very high respectively.
In \hyperref[fig:figure1]{Figure 1(c)} 
the ensemble diverges more than the single model. As the retention fraction increases, the ensemble shows a greater divergence from the single model. From this observation we infer that the ensemble is comparatively more responsive to distributional shift than the single model. MWSE assesses the sensitivity of a model to distribution shift by using error and uncertainty in conjunction with one another. The MWSE-retention curve when studied in tandem with the MSE-retention curve gives more insight into the quality of estimated uncertainty.  

\section{Conclusion and Future Work}
\label{others}
This work presents three primary metrics (RMSE/RMV ratio, LENCE and LL-FU) to quantify and evaluate predictive uncertainty under dataset shift. We used the Shifts Weather Prediction Dataset\cite{malinin2021shifts} for validation of the proposed metrics. Through these metrics we aimed to jointly evaluate predictive uncertainty and robustness of a model and its performance under dataset shift. While we focused only on regression tasks, we believe there is potential for these metrics to be extrapolated to classification as well as machine translation tasks. Another avenue of future research can be to incorporate the memory and computational efficiency of different methods into the evaluation metrics. We hope our work helps the community and inspires further research on evaluation metrics for shifted multi-modal data.  
\begin{ack}
We would like to thank \emph{Mars Rover Manipal} for providing the necessary resources for our research.
\end{ack}

\bibliography{References}


\newpage

\end{document}